\pdfoutput=1

\documentclass[11pt]{article}

\usepackage[preprint]{acl}

\usepackage{times}
\usepackage{latexsym}
\usepackage{multicol}
\usepackage{multirow, makecell, caption}
\usepackage{colortbl}
\usepackage{tikz}
\usepackage{arydshln}
\usepackage{pgfplots}
\usepackage{amsmath}
\usepackage{amssymb}
\usepackage{graphicx}
\usepackage{tabularx}
\usepackage{enumerate}
\usepackage{tcolorbox}
\usepackage{arydshln}
\usepackage{booktabs}
\usepackage{inconsolata}
\usepackage{microtype}
\usepackage{xcolor}
\usepackage{algorithm}
\usepackage{algorithmic}
\usepackage[utf8]{inputenc}
\usepackage{microtype}
\usepackage{inconsolata}
\usepackage{graphicx}
\usepackage{pifont}
\usepackage{xspace}
\usepackage{tcolorbox} 
\tcbuselibrary{listingsutf8}
\usepackage{listings}
\usepackage{hyperref}
\usepackage{enumitem}

\title{The Price of Format: Diversity Collapse in LLMs}

\author{Longfei Yun, Chenyang An, Zilong Wang, Letian Peng$^*$, Jingbo Shang\thanks{$\ $  Corresponding authors. } \\
University of California, San Diego \\
  \texttt{\{loyun, c5an, ziw049, lepeng, jshang\}@ucsd.edu}
  }

\begin{document}
\maketitle
\begin{abstract}
Instruction-tuned large language models (LLMs) employ structured templates, such as role markers and special tokens, to enforce format consistency during inference. However, we identify a critical limitation of such formatting: it induces a phenomenon we term diversity collapse, where the model generates semantically similar outputs for open-ended inputs, undermining creativity and variability. We systematically evaluate this effect across tasks like story completion and free-form generation, finding that (1) diversity collapse persists even under high-temperature sampling, and (2) structural tokens in templates significantly constrain the model’s output space. To contextualize these findings, we fine-tune the same model using a range of structured prompts and then evaluate them across three axes: downstream task performance, alignment behavior, and output diversity.
Our analysis shows that format consistency between fine-tuning and inference is crucial for structure-sensitive tasks (e.g., GSM8K, IFEval), but has marginal influence on knowledge-heavy tasks (e.g., MMLU, WebQuestions). In contrast, output diversity is primarily governed by the presence or absence of structural tokens, with minimal formatting yielding the most diverse outputs.
These findings reveal that current prompting conventions, while beneficial for alignment, may inadvertently suppress output diversity, underscoring the need for diversity-aware prompt design and instruction tuning.\footnote{Code is available at \url{https://github.com/LongfeiYun17/diversityCollapse}.}
\end{abstract}

\section{Introduction}
Instruction-tuned LLMs commonly adopt structured prompt templates that include role markers such as \texttt{<|user|>} and \texttt{<|assistant|>}, as well as special tokens like \texttt{<|begin\_of\_text|>}. These templates organize inputs and outputs into a dialogue-style format and are widely used in both open-source~\citep{dubey2024llama, qwen3} and proprietary models~\citep{achiam2023gpt, anil2023gemini}.

\begin{figure}[t]
    \centering        
    \includegraphics[width=0.50\textwidth]{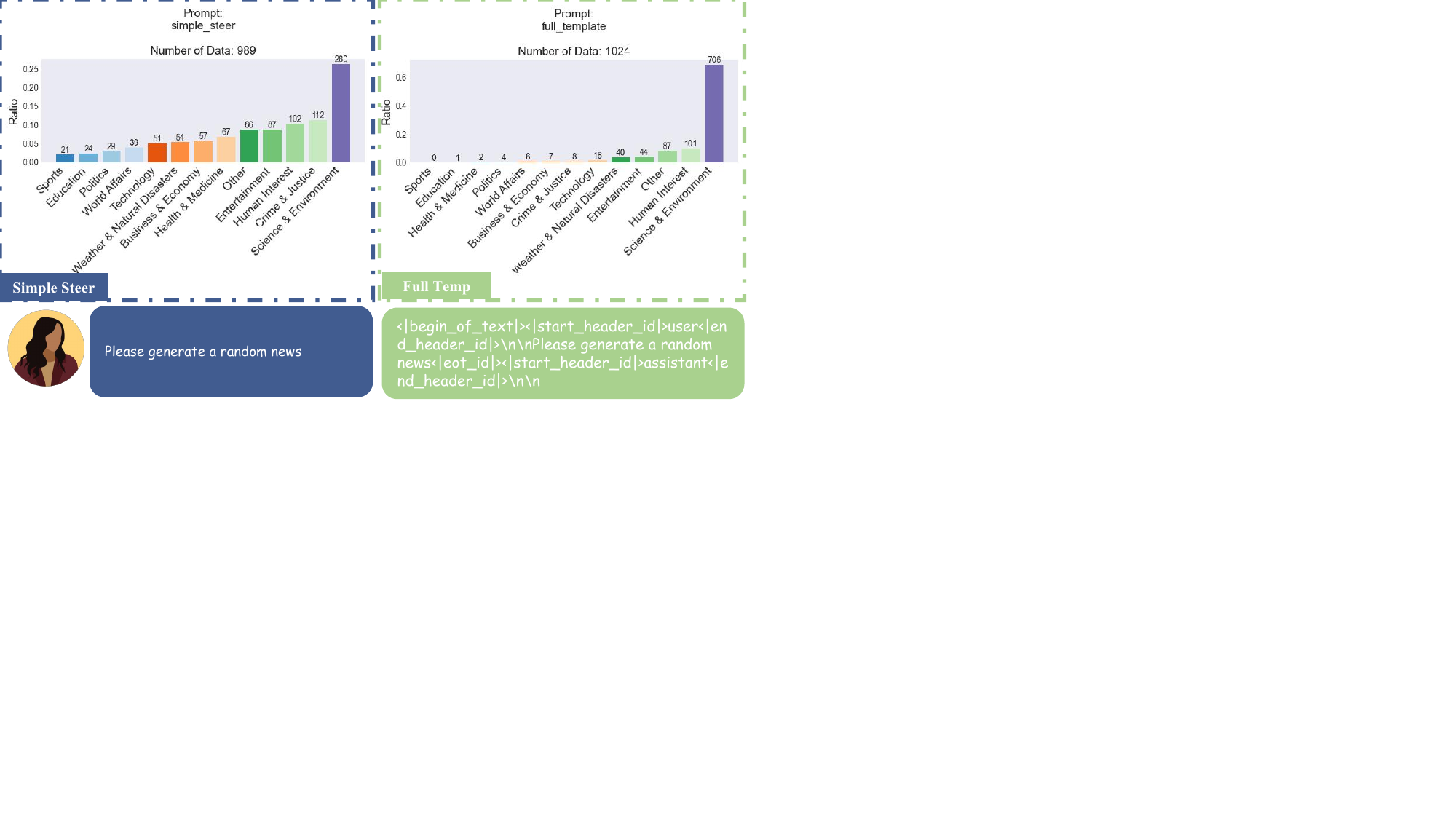}
    \caption{News generation results under simple prompt (Left) and full chat template prompt (Right). Templated prompting significantly reduces topic diversity.}
    \vspace{-1em}
    \label{fig:teaser}
\end{figure}

\noindent However, while such formatting improves consistency and alignment, we find that it significantly reduces the diversity of model outputs in an open-ended generation. As shown in \autoref{fig:teaser}, we generate 1024 news headlines using the same instruction (\textit{Please generate a random news}) under two prompting strategies, and classify the topics into predefined categories. The simple prompt yields a broad topic distribution across domains such as sports, health, and politics, whereas the templated prompt produces overwhelmingly Science-related content, indicating a sharp drop in topical diversity.
This effect persists even under high-temperature decoding, suggesting that the loss of diversity stems not from decoding randomness but from the template structure itself. These findings complement and extend prior work showing that instruction tuning can reduce output variability due to data distribution skew~\citep{mccoy2023embers}, training objectives~\citep{li2024entropic}, and alignment pressures~\citep{omahony2024attributing, kim2024knowledge}.

\noindent  We hypothesize that instruction-tuned models exposed to repeated structural templates during instruction tuning may internalize these patterns as strong generation priors, leading to overly deterministic or repetitive outputs in response to open-ended inputs. To validate this hypothesis, we empirically investigate the effect across multiple instruction-tuned LLMs and evaluate them on a suite of creative generation tasks. Our results show that structured prompts consistently yield lower semantic and topical diversity than simple prompts.

\noindent To pinpoint the source of diversity collapse, we conduct a controlled set of prompt-format ablations that gradually remove structural elements such as system tokens, role markers (e.g., \texttt{<|user|>}), and dialogue formatting. These prompting strategies are selected to represent a progression from highly structured, alignment-driven formats to fully open-ended instructions, enabling us to isolate the impact of each structural component.
We find that removing or replacing special tokens leads to only modest improvements in diversity. Even prompts with plain-text role indicators continue to limit diversity. In contrast, prompts presented as simple task instructions without any formatting achieve the highest diversity across models and tasks. This suggests that structural cues, even when lightweight, preserve behavioral priors from instruction tuning. Fully structure-free prompting remains the most effective way to recover expressive variation. To further understand the behavioral constraints imposed by templates, we also analyze output entropy across decoding steps (\S\ref{sec:ablations}) and find that structured prompts lead to lower entropy early in generation. This suggests that structural tokens act as behavioral anchors, causing the model to commit prematurely to narrow trajectories.

\noindent We next study how the impact of instruction tuning varies across tasks. To this end, we perform instruction tuning using different prompt strategies on the same dataset. We find that prompt formatting is essential for structure-sensitive benchmarks such as IFEval~\citep{zhou2023instruction} and GSM8K~\citep{cobbe2021training}, but can hinder performance on knowledge-intensive tasks. Overall, response quality is primarily determined by the consistency between tuning and inference, while the specific prompt format used at inference time plays a comparatively minor role.

\noindent Our contributions are as follows:

\begin{itemize}
\item We identify and empirically demonstrate a significant diversity collapse effect in an open-ended generation when using structured prompt templates. 
\item We conduct prompt ablations to isolate the effect of structural elements, and show that even lightweight formatting induces anchoring effects that reduce semantic diversity.
\item We analyze decoding dynamics and find that structured prompts suppress early-stage entropy, indicating strong anchoring effects learned during instruction tuning.
\item We find that prompt formatting is important for structure-sensitive tasks (e.g., IFEval, GSM8K) but can hurt performance on knowledge-intensive ones; overall, response quality depends more on the consistency between instruction tuning and inference-time prompting than on the presence of formatting templates.
\end{itemize}
\section{Related Work}
\subsection{Format Following}
The study of LLMs' capability to follow instructions was initially tackled by IFEval~\cite{zhou2023instruction}. INFOBENCH~\cite{qin2024infobench} expanded on this by covering a wider range of instructions. FOFO~\cite{xia2024fofo} is a benchmark dedicated entirely to evaluating LLMs' ability to follow format constraints. 
UltraBench~\cite{yun2025ultragen} explores LLMs' abilities under an extreme number of constraints. However, these studies do not investigate whether format instructions impact downstream task performance.

\subsection{Post-Training and Diversity Collapse} 
While instruction tuning and RLHF have significantly enhanced the reliability and helpfulness of LLMs, several studies have raised concerns about their unintended effects on output diversity. 
\citet{bai2022constitutional, ouyang2022training} first identified the so-called \textit{alignment tax}, where models exhibit diminished in-context learning abilities following Reinforcement Learning from Human Feedback (RLHF). 
Subsequent work by \citet{kirk2023understanding} further demonstrated that RLHF reduces output diversity, highlighting a tendency toward overfitting. Notably, these effects are not unique to RLHF: similar limitations arise under Supervised Fine-Tuning (SFT) alone, as shown by \citet{ouyang2022training, omahony2024attributing}. 
\citet{turpin2023language} formalized \textit{mode collapse} in instruction-tuned LLMs, noting sharp reductions in entropy and increased answer determinism. 
\citet{li2024entropic} further shows that the cross-entropy loss maximizes the likelihood of observed data without accounting for alternative plausible outputs, thereby contributing to reduced generative diversity.
\section{Experiment Setting}
\begin{table*}[htbp]
    \centering
    \begin{minipage}{\textwidth}
        \centering
        \newcolumntype{g}{>{\columncolor{green!10}}c}
        \renewcommand{\arraystretch}{1.22}
        \small
        \resizebox{\textwidth}{!}
        {
        \begin{tabular}{llccccccccc}
        \toprule
        \multirow{3}{*}{Model} & \multirow{3}{*}{Prompt Mode} & 
        \multicolumn{3}{c}{\textbf{Commonsense}} ($\uparrow$) & 
        \multicolumn{3}{c}{\textbf{Story Completion}} ($\uparrow$) &
        \multicolumn{3}{c}{\textbf{Open-ended Generation}} ($\uparrow$) \\
        \cmidrule(lr){3-5} \cmidrule(lr){6-8} \cmidrule{9-11} 
                & & CommonGen & ELI5 & NQ & WritingPrompts & ROCStory & Story\_Cloze & News & Travel & Books \\
        \midrule
        \multirow{2}{*}{Llama-3-8B-Instruct} 
        & Full Template & \textbf{0.2884} & 0.1438 & 0.1556 & 0.3278 & 0.1922 & 0.2015 & 0.0538 & 1.3098 & 1.4881 \\
        &  Simple Steer & 0.2692 & \textbf{0.2091} & \textbf{0.2115} & \textbf{0.4195} & \textbf{0.3845} & \textbf{0.3792} & \textbf{0.1399} & \textbf{2.5029} & \textbf{4.0250} \\
        \midrule
        
        \multirow{2}{*}{Tulu-3-8B-SFT} 
        & Full Template & 0.3200 & 0.4135 & 0.4250 & 0.5239 & 0.3920 & 0.3977 & 0.1706 & \textbf{3.8673} & 4.4450 \\
        &  Simple Steer & \textbf{0.3627} & \textbf{0.4958} & \textbf{0.4746} & \textbf{0.6160} & \textbf{0.5553} & \textbf{0.5119} & \textbf{0.2185} & 3.7306 & \textbf{4.8256} \\
        \midrule

        \multirow{2}{*}{Qwen2.5-7B-Instruct} 
        & Full Template & 0.1838 & 0.1178 & 0.1196 & 0.3133 & 0.1760 & 0.1786 & \textbf{0.1200} & 1.0215 & \textbf{4.2973} \\
        &  Simple Steer & \textbf{0.2390} & \textbf{0.2357} & \textbf{0.2152} & \textbf{0.4293} & \textbf{0.3744} & \textbf{0.3701} & 0.1090 & \textbf{3.3677} & 4.0948 \\
        \midrule

        \multirow{2}{*}{Mistral-7B-Instruct-v0.1} 
        & Full Template & 0.2884 & 0.1856 & 0.2241 & 0.3696 & 0.2106 & 0.1938 & 0.1037 & 2.5062 & 2.2940 \\
        & Simple Steer & \textbf{0.3020} & \textbf{0.2872} & \textbf{0.4280} & \textbf{0.4821} & \textbf{0.4667} & \textbf{0.3879} & \textbf{0.1492} & \textbf{2.8922} & \textbf{3.1499} \\
        \midrule

        \multirow{2}{*}{Phi-3.5-mini-instruct} 
        & Full Template & 0.2616 & 0.1536 & 0.1900 & 0.3551 & 0.2602 & 0.2528 & 0.0734 & 1.6466 & 2.6115 \\
        &  Simple Steer & \textbf{0.2921} & \textbf{0.2127} & \textbf{0.3171} & \textbf{0.4558} & \textbf{0.3721} & \textbf{0.3671} & \textbf{0.1533} & \textbf{3.2307} & \textbf{4.3440} \\
        \bottomrule
        \end{tabular}
        }
        \caption{Performance comparison of instruction-tuned language models on nine tasks under two prompting conditions: Full Template and Simple Steer. Simple Steer consistently yields higher diversity than Full Template.
        }
        \label{tab:diversity_collapse}
    \end{minipage}
\end{table*}

\paragraph{Problem Formulation}
We define the \textit{diversity score} as a quantitative measure of variation in model outputs. Following the evaluation metrics in~\S\ref{para:eval_metrics}, we use either the average semantic distance between sentence embeddings~\citep{tevet-berant-2021-evaluating,han-etal-2022-measuring} or the entropy~\citep{thoughtworks2025semanticentropy, chen2024synthetic} over generated topics to evaluate diversity. Embedding-based metrics capture semantic variation between outputs, while entropy-based metrics reflect topic coverage across generations. Together, they offer complementary views of diversity.

\noindent Based on this, we compute two diversity scores for each task: 
$D_{\text{simple}}$, the average diversity score under the simple prompt condition, and 
$D_{\text{template}}$, the average diversity score under the full chat template condition.

\noindent We define \textit{diversity collapse} as the phenomenon where
\[
D_{\text{template}} \ll D_{\text{simple}}
\]
That is, diversity under the template prompting mode is significantly lower than that under the simple prompting mode, even when using high-temperature decoding.
\paragraph{Target Models} 
We select five instruction-tuned models with varying architectures and alignment strategies: 
(1) Llama-3-8B-Instruct~\citep{llama3modelcard}, 
(2) Tulu-3-8B-SFT~\citep{lambert2024tulu3}, 
(3) Qwen2.5-7B-Instruct~\citep{qwen2.5}, 
(4) Mistral-7B-Instruct-v0.1~\citep{jiang2024identifying}, and 
(5) Phi-3.5-mini-instruct~\citep{abdin2024phi}. 
This set enables a robust evaluation of whether diversity collapse persists across different model design choices.

\paragraph{Tasks} 
We evaluate on nine tasks spanning commonsense reasoning~\citep{lin2019commongen, fan-etal-2019-eli5, kwiatkowski2019natural}, story completion~\citep{fanHierarchical2018, mostafazadeh2016corpus, mostafazadeh-etal-2017-lsdsem}, and preference modeling. This diverse task set allows us to examine whether diversity collapse is a universal issue across task types.

\begin{enumerate}[itemsep=0.1pt, topsep=0.5pt, parsep=0pt, partopsep=0pt]
\item \textbf{Commonsense}: The model is given either a question (e.g., \textit{How do muscles grow?}) or a set of concepts (e.g., \textit{hay, eat, horse}) and is asked to generate a plausible, commonsense-based response.

\item \textbf{Story Completion}: The model is provided with an opening prompt (e.g., \textit{The moon is actually a giant egg, and it has just started to hatch.}) and tasked with completing the story in a coherent and creative manner.

\item \textbf{Open-ended Generation}: We assess the diversity of generated content by computing the entropy of entities (e.g., topics, locations, or titles) mentioned in the model outputs. For instance, in the news generation task, we measure topic diversity by analyzing the distribution of topics across generated articles.
\end{enumerate}

\noindent For the Commonsense and Story Completion tasks, we randomly sample 512 prompts from the test split of each dataset and generate 10 responses per prompt. For the Open-ended Generation tasks, we generate 1,024 responses in total. All generations are performed using temperature $T{=}1.0$ and top-$p{=}0.9$ sampling.

\paragraph{Evaluation Metrics} 
\label{para:eval_metrics}
We report semantic diversity for the Commonsense and Story Completion tasks using sentence embedding distances, following prior work~\citep{tevet-berant-2021-evaluating}, which found embedding-based metrics to be more effective than n-gram-based alternatives. For the Open-ended Generation tasks, we compute the entropy of extracted entities to assess label-level diversity. Traditional metrics such as distinct-n and self-BLEU are reported in~\autoref{appendix:traditional_metrics}.

\begin{enumerate}
    \item \textbf{Semantic Diversity}: We measure the average pairwise distance between sentence embeddings of responses to the same prompt. Given $N$ prompts, each with $k$ responses, we compute sentence embeddings using the all-MiniLM-L6-v2 model~\citep{reimers-2020-multilingual-sentence-bert}. The overall diversity score is:
    \[
    D = \operatorname{avg}_{n} \left( \operatorname{avg}_{i < j} \left( 1 - \cos\left(\mathbf{e}_i^{(n)}, \mathbf{e}_j^{(n)} \right) \right) \right)
    \]

    where $\mathbf{e}_i^{(n)}$ is the embedding of the $i$-th response to the $n$-th prompt.

    \item \textbf{Label Diversity}: For each of the $N$ generations $\{y_i\}_{i=1}^N$, we use GPT-4o~\citep{achiam2023gpt} to extract a single entity label $e_i$. Let $P(e)$ denote the empirical distribution over the label set $\mathcal{E} = \{e_i\}$. The topic diversity is computed as the normalized entropy:
    \[
    D_{\text{topic}} = \frac{-\sum_{e \in \mathcal{E}} P(e) \log P(e)}{\log |\mathcal{E}|}.
    \]
\end{enumerate}

\section{Understanding Diversity Collapse}

We begin by examining how diversity collapse presents in model outputs (\S\ref{sec:diversity-metrics}), and follow with an analysis of its underlying causes (\S\ref{sec:ablations}).

\subsection{Templates Reduce Output Diversity}
\label{sec:diversity-metrics}

To test whether prompt templates reduce output diversity, we compare a natural, minimal prompt format (\textit{simple steer}) with the standard full chat-style template (\textit{full template}) (See ~\autoref{tab:prompt_templates_all_models}). For each model, we fix the SFT data and vary only the inference-time prompt format to isolate the effect of prompt structure.

\begin{figure}[htbp]
    \centering        \includegraphics[width=0.49\textwidth]{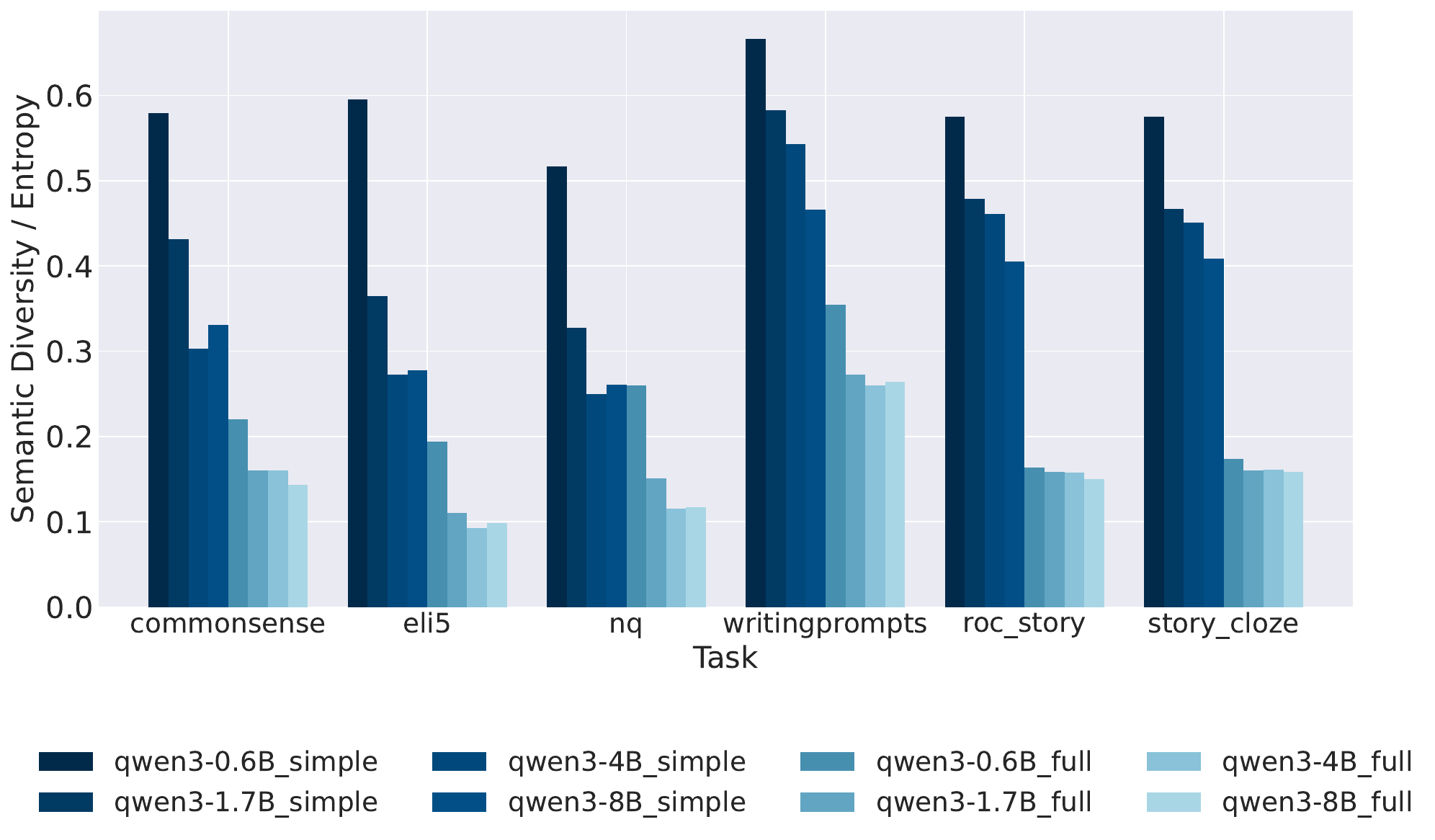}
    \caption{Semantic diversity comparison across Qwen3~\citep{qwen3} model sizes under two prompting modes, excluding the thinking mode. The results show that diversity collapse occurs consistently across model scales.}
    \label{fig:results_across_models}
\end{figure}

\noindent As shown in~\autoref{tab:diversity_collapse}, we observe a consistent pattern across all models and task types: \textbf{full chat templates significantly reduce output diversity compared to simple steer prompts.}
 The bar chart in~\autoref{fig:results_across_models} further confirms this trend: across all model sizes, simple steer prompts consistently yield higher semantic diversity. This gap persists at larger model scales, suggesting that template-induced diversity collapse is not mitigated by increased capacity.

\begin{figure*}[tbp]
    \centering        \includegraphics[width=0.98\textwidth]{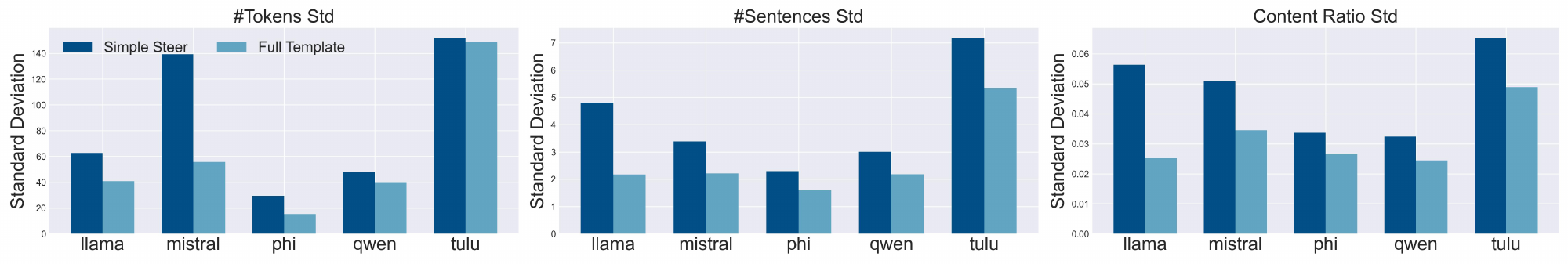}
    \caption{Structural diversity across prompting modes in the news generation task, measured by the standard deviation of content word ratio (left), sentence count (middle), and token length (right). }
    \label{fig:structure_analysis}
\end{figure*}

\noindent We also assess structural diversity by computing the standard deviation of the token count, sentence count, and content word ratio\footnote{\#content words / \#total words, where content words exclude stopwords}. As shown in~\autoref{fig:structure_analysis}, simple steer prompts consistently lead to greater structural variation than full templates across all models. This suggests that chat templates constrain not only what models say, but also how they say it, reducing variability in form and content, and thereby narrowing the expressive space.

\subsection{Dissecting Template-Induced Collapse}
\label{sec:ablations}

\paragraph{Chat Templates as Behavioral Triggers}
To systematically examine how prompt structure affects output diversity, we evaluate four prompting strategies with varying degrees of structural complexity. An example for LLaMA is shown in~\autoref{tab:prompt_template_example}.

\begin{enumerate}[itemsep=0.1pt, topsep=0.5pt, parsep=0pt, partopsep=0pt]
    \item \textbf{Full Template}: Standard chat-style format with system/user/assistant tokens, closely aligned with the conventions used during instruction tuning.

    \item \textbf{Fake Template}: Retains the structural layout of chat prompts but replaces standard tokens with semantically meaningless variants.

    \item \textbf{Minimal Dialog}: Removes all special tokens while retaining plain-text role indicators (e.g., \textit{User}, \textit{Assistant}) to preserve natural dialogue flow.

    \item \textbf{Simple Steer}: A minimal, structure-free prompt containing only the task description.
\end{enumerate}


\begin{table}[htbp]
    \centering
    \tiny
    \renewcommand{\arraystretch}{1.3}
    \begin{tabular}{p{1.3cm}|p{5.7cm}}
        \toprule
        \textbf{Prompt Mode} & \textbf{LLaMA Prompt Example} \\
        \midrule
        \texttt{full\_template} & \texttt{\detokenize{<|begin_of_text|><|start_header_id|>user<|end_header_id|>}} \newline 
        \texttt{\detokenize{[instruction]}} \newline 
        \texttt{\detokenize{<|eot_id|><|start_header_id|>assistant<|end_header_id|>}} \\
        
        \texttt{fake\_template} & \texttt{\detokenize{<#init_text#><#random_header#>user<#/random_header#>}} \newline 
        \texttt{\detokenize{[instruction]}} \newline 
        \texttt{\detokenize{<#eod#><#random_header#>assistant<#/random_header#>}} \\
        
        \texttt{minimum\_dialog} & \texttt{\detokenize{user}} \newline 
        \texttt{\detokenize{[instruction]}} \newline 
        \texttt{\detokenize{assistant:}} \\
        
        \texttt{simple\_steer} & \texttt{\detokenize{[instruction]}} \\
        
        \bottomrule 
    \end{tabular}
    \caption{Prompt formats for different modes used with LLaMA. Here, \texttt{[instruction]} is the task-specific input.}
    \label{tab:prompt_template_example}
\end{table}

\begin{figure}[h]
    \centering        \includegraphics[width=0.48\textwidth]{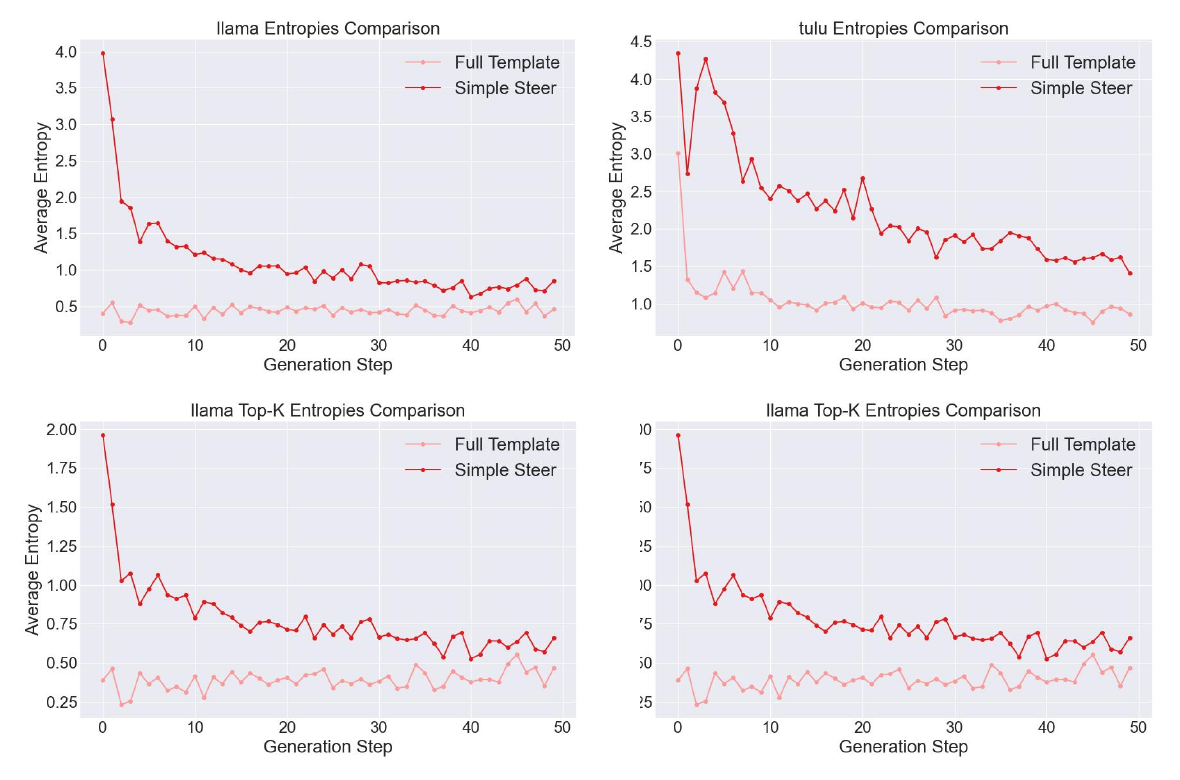}
    \caption{Entropy of the output space across decoding steps with and without templates. The figure shows that using a template significantly reduces entropy, indicating a more constrained and predictable output distribution.}
    \label{fig:output_entropy}
\end{figure}

\begin{figure*}[htbp]
    \centering        \includegraphics[width=0.98\textwidth]{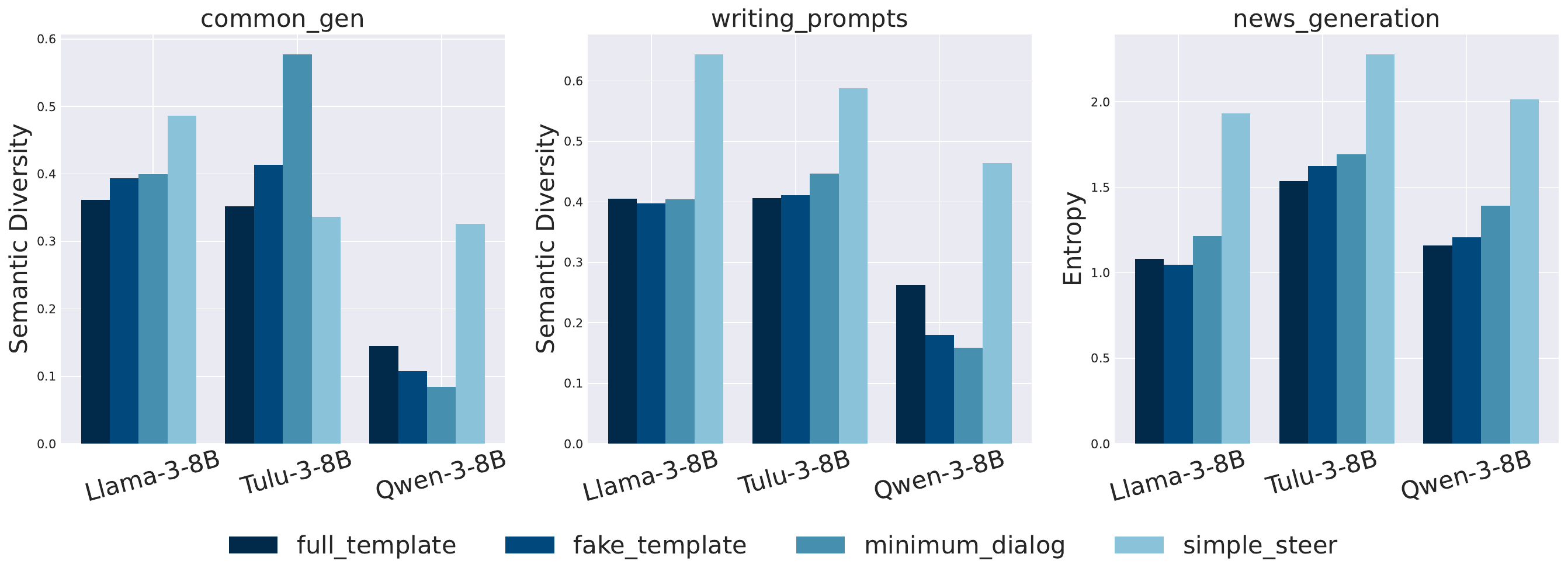}
    \caption{Performance comparison across prompting modes (\textbf{Full Template}, \textbf{Fake Template}, \textbf{Minimum Dialog}, and \textbf{Simple Steer}) for three instruction-tuned language models on three representative tasks.}
    \label{fig:fine_grained_bar.pdf}
\end{figure*}

\noindent Our results in~\autoref{fig:fine_grained_bar.pdf} highlight how different prompt structures affect output diversity:
\begin{enumerate} [itemsep=0.1pt, topsep=0.5pt, parsep=0pt, partopsep=0pt]
    \item \textbf{Simple Steer yields the highest diversity:} Prompts without any structural framing consistently produce the highest semantic diversity and entropy, confirming that fully removing template structure is the most effective strategy.

    \item \textbf{Even minimal structure reduces diversity:} Although Minimal Dialog prompts outperform Full and Fake Templates, they still yield substantially lower diversity than Simple Steer. This suggests that even lightweight structural cues, such as plain-text role markers can constrain generative variation.
\end{enumerate}

\noindent Taken together, these findings indicate that diversity collapse is driven not only by rigid templates, but by structural conventions more broadly. Only fully structure-free prompting reliably restores expressive flexibility.

\paragraph{Chat Templates Narrow the Output Space}
To quantify how structured prompts affect generation dynamics, we measure token-level entropy at each decoding step. Specifically, we sample 128 instructions and track the entropy of the model’s output distribution over 50 generation steps. As shown in~\autoref{fig:output_entropy}, chat-style prompts consistently produce lower entropy than simple steer prompts. This suggests that chat templates constrain the model’s output space, resulting in more deterministic and less varied generations.
\section{Mitigation Strategies and Further Analysis}
We answer these questions in this section: 
\begin{enumerate} [itemsep=0.1pt, topsep=0.5pt, parsep=0pt, partopsep=0pt]
    \item Can structural prompt variants mitigate diversity collapse? (\S\ref{subsec:experiment_setup}, \S\ref{subsec:main_results})
    \item How do diversity-preserving prompts affect downstream task performance and instruction-following ability? (\S\ref{subsec:downstream}) 
    \item Can higher decoding temperatures (\S\ref{subsec:effect_of_temperature}) or explicit prompting for creativity restore lost diversity (\S\ref{subsec:explicit_prompts})?
\end{enumerate}

\begin{table*}[htbp]
    \centering
    \small
    \renewcommand{\arraystretch}{1.22}
    \resizebox{\textwidth}{!}{
    \begin{tabular}{lccccccccc}
    \toprule
    \multirow{3}*{Method} & 
    \multicolumn{3}{c}{\textbf{Commonsense} ($\uparrow$)} & 
    \multicolumn{3}{c}{\textbf{Story Completion} ($\uparrow$)} &
    \multicolumn{3}{c}{\textbf{Open-ended Generation} ($\uparrow$)} \\
    \cmidrule(lr){2-4} \cmidrule(lr){5-7} \cmidrule(lr){8-10}
    & CommonGen & ELI5 & NQ & WritingPrompts & ROCStory & Story\_Cloze & News & Travel & Books \\
    \midrule
    Full Template & 0.3717 & 0.4622 & 0.4992 & 0.4249 & 0.3648 & 0.3765 & 1.5689 & 4.2083 & \textbf{4.4126} \\
    Simple Steer & \textbf{0.4474} & \underline{0.5238} & \underline{0.6666} & \textbf{0.5513} & \textbf{0.4740} & \textbf{0.4563} & \textbf{2.3552} & 3.5435 & 3.6572 \\
    Mixed Template & 0.3175 & 0.4516 & 0.4487 & 0.4302 & 0.3673 & 0.3762 & 1.9646 & \underline{4.7086} & \underline{4.1715} \\
    Natrual Instruction & \underline{0.4227} & \textbf{0.5850} & \textbf{0.7162} & \underline{0.5263} & \underline{0.3983} & 0.3911 & \underline{2.2312} & \textbf{4.7115} & 3.9262 \\
    Mixed Training & 0.3970 & 0.4635 & 0.5067 & 0.4337 & 0.3846 & \underline{0.4074} & 1.9853 & 4.3866 & 3.8618 \\
    \bottomrule
    \end{tabular}
    }
    \caption{Diversity scores ($\uparrow$) on commonsense reasoning, story completion, and open-ended generation tasks. The best value in each column is in bold, and the second-best is underlined. Removing structural formatting (Simple Steer, Natural Instruction) generally improves output diversity.}
    \label{tab:different_prompt_strategies}
    \vspace{-1em}
\end{table*}

\subsection{Experiment Setup}
\label{subsec:experiment_setup}
\paragraph{Dataset} 
We use the \textsc{tulu-3-sft-mixture} dataset~\citep{lambert2024tulu3} due to its broad coverage of core instruction-following capabilities. Curated from high-quality public and persona-driven sources~\citep{ge2024scaling}, it emphasizes data diversity, quality, and licensing compliance. The dataset has also undergone rigorous decontamination to ensure fair evaluation.

\paragraph{Training Settings}
We fine-tune a \textsc{LLaMA-3.2-3B} model~\citep{grattafiori2024llama} for three epochs using a batch size of 8, a sequence length of 1024, and a learning rate of $6 \times 10^{-6}$.

\paragraph{Baselines}
We investigate whether modifying prompt structure can mitigate diversity collapse without compromising generation quality. To this end, we evaluate five prompting strategies with varying levels of structural complexity.

\begin{enumerate} [itemsep=0.1pt, topsep=0.5pt, parsep=0pt, partopsep=0pt]
    \item \textbf{Full Template}: Uses the standard chat-style prompt format adopted during instruction tuning, including special tokens and explicit role markers.

    \item \textbf{Simple Steer}: Uses the same instruction-tuned checkpoint as Full Template, but provides only the core instruction at inference time, with no structural formatting.

    \item \textbf{Mixed Template}: Randomly selects one prompt format from a pool of instruction-tuned templates to introduce format-level variation during inference.

    \item \textbf{Natural Instruction}: Uses the same prompt format as Simple Steer (structure-free), but the model is fine-tuned directly on natural instructions without any chat-style formatting.

    \item \textbf{Mixed Training}: Augments the instruction-tuning data with pretraining-style samples that include no prompt formatting, comprising one-third of the training corpus.
\end{enumerate}

\subsection{Main Results}
\label{subsec:main_results}
\paragraph{Homogeneity Drives Diversity Collapse}
Although the Mixed Template setting introduces format variation, it consistently underperforms compared to other prompting strategies across nearly all tasks. Our comparison between Full Template and Mixed Template shows that increasing the number of formats does not meaningfully improve output diversity. This suggests that the collapse arises not from overuse of a single template, but from structural similarities shared across templates. All five variants follow the same chat-style pattern with explicit role markers and turn-taking, which is sufficient to constrain the model's generative behavior.
These results suggest that structural homogeneity, rather than format repetition alone, maybe a key factor contributing to diversity collapse.

\paragraph{Natural Instruction Matches Simple Steer}
The Natural Instruction setting, which removes all special tokens and role markers, performs on par with Simple Steer in diversity across tasks. This suggests that in the absence of structural triggers such as special tokens or role indicators, the model’s expressive capacity is preserved.

\paragraph{Mixed Training Provides Limited Improvement}
We augment the SFT data with pretraining-style samples that contain no template structure. However, the Mixed Training model shows only marginal improvements over the Full Template baseline. This suggests that limited exposure to unstructured data during fine-tuning may be insufficient to counteract the behavioral priors induced by chat-style prompts at inference time.

\subsection{Downstream Performance and Instruction-Following Tradeoffs}
\label{subsec:downstream}
\begin{table}[htbp]
    \centering
    \begin{minipage}{.49\textwidth}
        \renewcommand{\arraystretch}{1.22}
        \resizebox{\textwidth}{!}
        {
        \begin{tabular}{lcccccc}
            \toprule
            \multirow{3}{*}{Method} & \multicolumn{6}{c}{\textbf{Downstream Tasks}} \\
            \cmidrule(lr){2-7}
            & MMLU & GSM8K & HumanEval & WebQS & IFEval & WSC273\\
            \midrule
            Base Model & 0.5412 & 0.2623 & 0.2561 & 0.0915 & 0.1799 & 0.8168 \\
            Full Template   & 0.4870 & 0.3935 & 0.1646 & 0.0349 & 0.3087 & 0.7399 \\
            Simple Steer    & 0.4880 & 0.2388 & \textbf{0.3048} & \textbf{0.0890} & 0.1645 & \underline{0.7912} \\
            Natural Instruction & 0.5104 & 0.4359 & 0.1280 & 0.0846 & 0.3142 & 0.7729 \\
            Mixed Training  & \underline{0.4912} & \underline{0.3972} & 0.1098 & 0.0492 & \underline{0.3179} & \textbf{0.8059}\\
            Mixed Template  & \textbf{0.5090 }& \textbf{0.4390} & \underline{0.2622} & \underline{0.0566} & \textbf{0.5336} & 0.7473 \\
            \bottomrule
        \end{tabular}
        }
    \end{minipage}
    \caption{We evaluate models on MMLU, GSM8K, HumanEval, WebQS, IFEval, and WSC273 to assess whether prompt formats impact real-world performance. \textbf{Bold} indicates the best score for each task, and \underline{underline} indicates the second-best.
    }
    \vspace{-1em}
    \label{tab:down_stream}
\end{table}

\subsubsection{Downstream Performance}
\noindent We evaluate each method on six downstream benchmarks to assess whether diversity-enhancing prompts affect task performance. The selected benchmarks span a broad range of capabilities: (1) multi-domain factual reasoning (MMLU~\citep{hendrycks2020measuring}), (2) mathematical problem solving (GSM8K~\citep{cobbe2021training}), (3) instruction following (IFEval~\citep{zhou2023instruction}), (4) factual question answering (WebQuestions~\citep{berant-etal-2013-semantic}), (5) structured code generation (HumanEval~\citep{chen2021evaluating}), and (6) commonsense pronoun resolution (WSC273~\citep{levesque2012winograd}). Evaluation details are provided in~\autoref{appendix:downstream_evaluation_details}.

\paragraph{Base Model Performance: Alignment May Hurt Knowledge Tasks}
The base model in the comparison reveals that instruction tuning and prompt templating do not universally improve downstream performance. On knowledge-intensive tasks such as MMLU and WebQuestions, the base model outperforms all instruction-tuned variants. This suggests that alignment procedures may inadvertently impair factual recall by overriding the model’s pretrained knowledge. In such cases, prompt formatting offers limited benefit, indicating that factual accuracy relies more on internal representations than on external scaffolding.

\paragraph{Format Consistency Benefits Certain Tasks}
We find that for structure-sensitive tasks such as GSM8K and IFEval, models achieve the best performance when the prompt format used at inference matches the format seen during fine-tuning. Both Full Template and Natural Instruction perform well on these tasks, each maintaining consistency between training and inference formats. In contrast, Simple Steer underperforms, likely due to its mismatch with the structured format used during training.  However, this pattern does not hold universally. On tasks such as HumanEval, where inputs already include rich syntactic signals (e.g., function headers, docstrings), Simple Steer outperforms other formats. This suggests that for tasks with strong intrinsic structure, additional prompting scaffolding may introduce noise rather than provide benefit. Other tasks, such as WSC273 and WebQS, show minimal sensitivity to prompt format.

\begin{figure}[htbp]
    \centering        \includegraphics[width=0.48\textwidth]{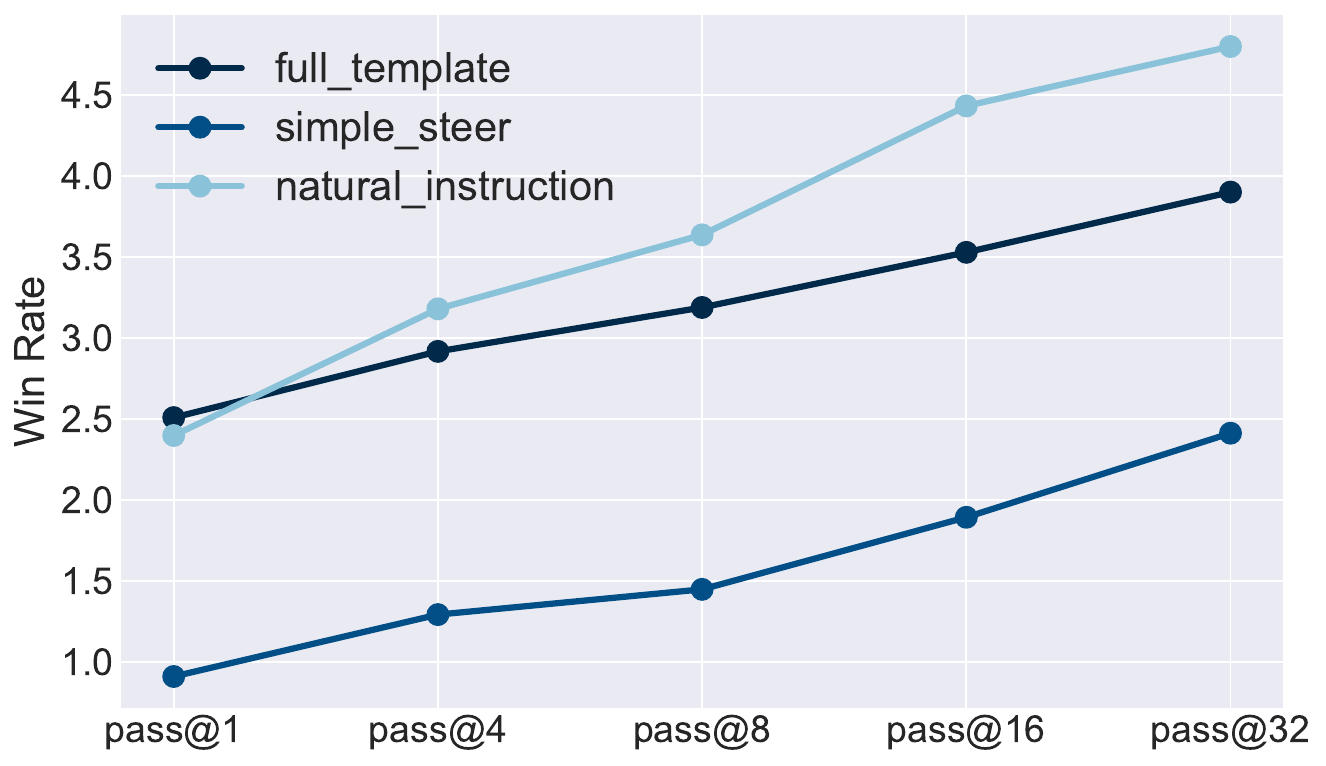}
    \caption{ Comparison of win rates on AlpacaEval under different prompting strategies. A reward model selects the best response among candidates. The win rate (y-axis) increases with the number of sampled responses (x-axis, pass@k), but full\_template consistently outperforms simple\_steer across all settings.}
    \label{fig:alpacaeval}
\end{figure}

\subsubsection{Instruction Tuning Enhance Response Quality} 
In this section, we demonstrate that although output diversity is affected by prompt structure, chat-style templates enhance the model’s ability to produce high-quality responses. We prompt each fine-tuned model to answer 805 questions from the \textsc{AlpacaEval} dataset~\citep{alpaca_eval, dubois2024length}. For each question, the model generates 32 responses, and a reward model selects the best one. We use \textsc{Skywork-Reward-Llama-3.1-8B-v0.2}\footnote{\url{https://huggingface.co/Skywork/Skywork-Reward-Llama-3.1-8B-v0.2}}, which achieves top performance on \textsc{RewardBench}~\citep{lambert2024rewardbench}.

\noindent These results suggest that while simple prompting improves generation diversity, it weakens the model’s ability to produce high-quality outputs, as measured by AlpacaEval. Interestingly, we find that the Natural Instruction setting achieves even better performance than the full instruction-tuning template. This suggests that instruction tuning is a key factor in enhancing response quality, even in the absence of special tokens or rigid chat formatting.

\begin{figure}[htbp]
    \centering        \includegraphics[width=0.48\textwidth]{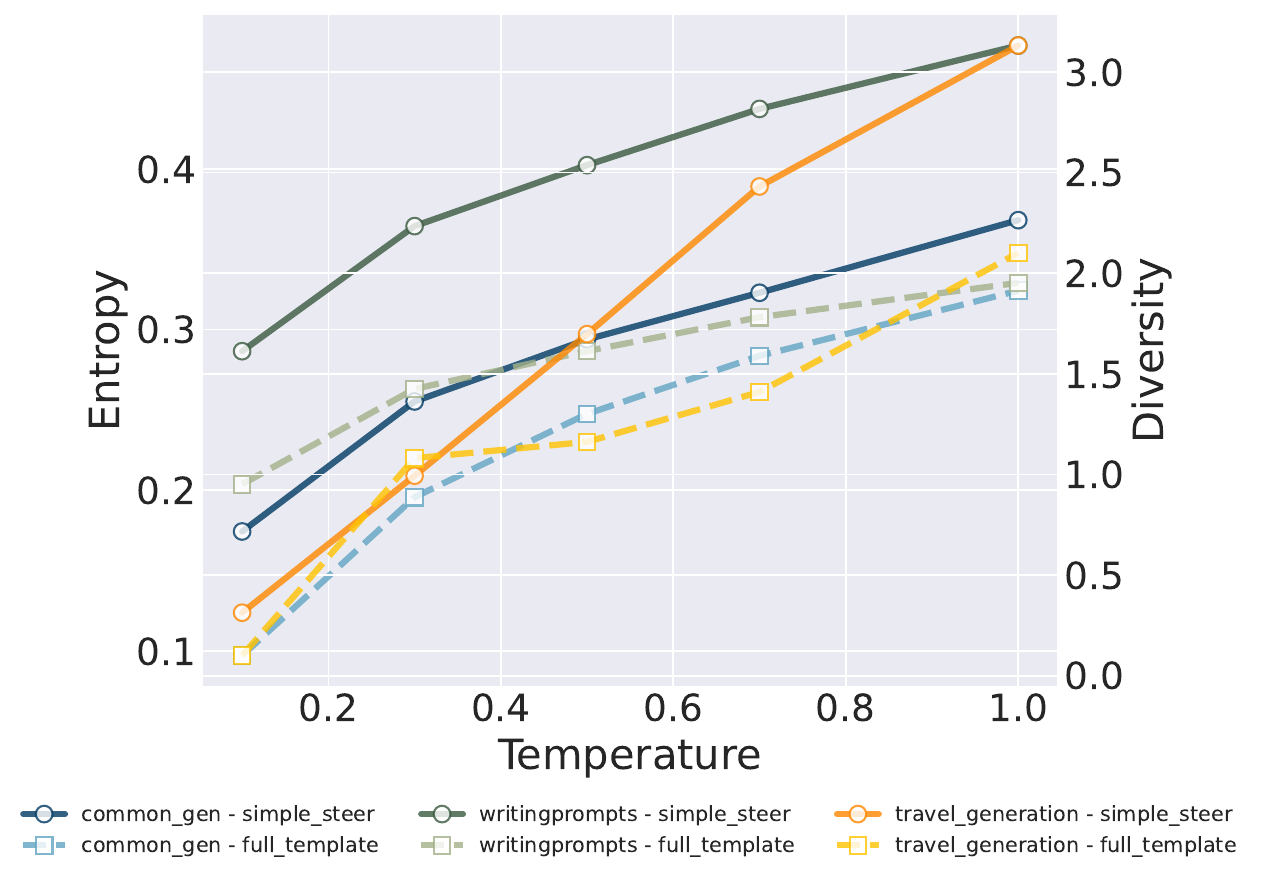}
    \caption{Effect of decoding temperature on semantic diversity and entropy across different generation tasks under simple steer and full template prompts. Higher temperature consistently increases diversity and entropy, while using a structured template (full template) notably limits this effect compared to simple steer prompts.}
    \label{fig:temperature}
    \vspace{-1em}
\end{figure}

\subsection{Effect of Decoding Temperature on Diversity}
\label{subsec:effect_of_temperature}
To further investigate the impact of decoding temperature on diversity collapse, we analyze how output diversity changes under different sampling temperatures. As shown in~\autoref{fig:temperature}, higher temperatures reliably increase both semantic diversity and entropy. However, prompts using full chat templates exhibit a muted response: their gains from temperature scaling are substantially smaller than those under simple steer prompts. This suggests that structured templates impose a significant constraint on the model’s generative freedom, limiting the benefits typically associated with higher-temperature decoding.

\subsection{Explicit Prompts for Diversity Still Fall Short}
\label{subsec:explicit_prompts}
\begin{table}[htbp]
\centering
\resizebox{0.49\textwidth}{!}{
\begin{tabular}{lccc}
\toprule
\textbf{Task} & \textbf{Full Template} & \textbf{Simple Steer} & \textbf{Full Template w/ Diversity} \\ 
\midrule
\multicolumn{4}{c}{\textit{Semantic Diversity}} \\
\midrule
CommonGen & 0.3242 & 0.3680 & \textcolor{green!60!black}{0.3668} \\
ELI5 & 0.1565 & 0.2998 & \textcolor{red}{0.2550} \\
Natural Questions (NQ) & 0.1879 & 0.3253 & \textcolor{red}{0.2570} \\
WritingPrompts & 0.3291 & 0.4767 & \textcolor{red}{0.3427} \\
ROCStory & 0.1986 & 0.3768 & \textcolor{red}{0.2264} \\
Story Cloze & 0.1980 & 0.3561 & \textcolor{red}{0.2261} \\
\midrule
\multicolumn{4}{c}{\textit{Entropy}} \\
\midrule
News Generation & 0.3186 & 1.6790 & \textcolor{red}{1.5763} \\
Travel Recommendation & 2.1002 & 3.1312 & \textcolor{green!60!black}{3.5087} \\
Book Recommendation & 2.1544 & 3.2290 & \textcolor{red}{0.9539} \\
\bottomrule
\end{tabular}}
\caption{Comparison of semantic diversity and entropy across tasks using different prompting methods with Llama-3b model. \textbf{Green} values indicate that explicitly prompting for diversity achieves comparable or better results than simple steer. \textbf{Red} values indicate limited improvement or remaining significantly below simple steer.}
\vspace{-1em}
\label{tab:prompt_effect}
\end{table}

\noindent In this section, we examine whether explicitly prompting the model to "be creative" can improve output diversity. Despite such encouragement within the chat template setting (Full Template w/ Diversity), the resulting diversity remains consistently lower than that achieved by minimal prompts (e.g., Simple Steer) across most tasks. As shown in~\autoref{tab:prompt_effect}, prompting for diversity does yield noticeable improvements over the default template in some cases (e.g., CommonGen, Travel Recommendation), but still fails to close the gap with Simple Steer. These results suggest that the structural constraints imposed by chat templates cannot be easily mitigated through surface-level prompting. 
\section{Conclusion}
In this paper, we identified and investigated the phenomenon of diversity collapse, revealing how structured chat templates significantly reduce semantic and topical diversity in instruction-tuned large language models. Through comprehensive empirical analyses and ablation studies, we demonstrated the structural factors causing this limitation and provided practical strategies to mitigate it. Our findings highlight the crucial role of prompt design in preserving model creativity, offering valuable insights for future research and applications of language models.
\section*{Limitations}
 First, our study compares only chat-style and simple-steer templates, so the observed diversity collapse may differ under other prompting strategies such as chain-of-thought or retrieval-augmented approaches. Second, we measure diversity using automated semantic and lexical metrics at the utterance level, leaving discourse-level variation and downstream impact for future work.

\bibliography{custom}

\begin{thebibliography}{43}
\providecommand{\natexlab}[1]{#1}

\bibitem[{Abdin et~al.(2024)Abdin, Aneja, Awadalla, Awadallah, Awan, Bach, Bahree, Bakhtiari, Bao, Behl et~al.}]{abdin2024phi}
Marah Abdin, Jyoti Aneja, Hany Awadalla, Ahmed Awadallah, Ammar~Ahmad Awan, Nguyen Bach, Amit Bahree, Arash Bakhtiari, Jianmin Bao, Harkirat Behl, et~al. 2024.
\newblock Phi-3 technical report: A highly capable language model locally on your phone.
\newblock \emph{arXiv preprint arXiv:2404.14219}.

\bibitem[{Achiam et~al.(2023)Achiam, Adler, Agarwal, Ahmad, Akkaya, Aleman, Almeida, Altenschmidt, Altman, Anadkat et~al.}]{achiam2023gpt}
Josh Achiam, Steven Adler, Sandhini Agarwal, Lama Ahmad, Ilge Akkaya, Florencia~Leoni Aleman, Diogo Almeida, Janko Altenschmidt, Sam Altman, Shyamal Anadkat, et~al. 2023.
\newblock Gpt-4 technical report.
\newblock \emph{arXiv preprint arXiv:2303.08774}.

\bibitem[{AI@Meta(2024)}]{llama3modelcard}
AI@Meta. 2024.
\newblock \href {https://github.com/meta-llama/llama3/blob/main/MODEL_CARD.md} {Llama 3 model card}.

\bibitem[{Anil et~al.(2023)}]{anil2023gemini}
Rohan Anil et~al. 2023.
\newblock \href {https://arxiv.org/abs/2312.11805} {Gemini: A family of highly capable multimodal models}.
\newblock \emph{arXiv preprint arXiv:2312.11805}.

\bibitem[{Bai et~al.(2022)Bai, Kadavath, Kundu, Askell, Kernion, Jones, Chen, Goldie, Mirhoseini, McKinnon et~al.}]{bai2022constitutional}
Yuntao Bai, Saurav Kadavath, Sandipan Kundu, Amanda Askell, Jackson Kernion, Andy Jones, Anna Chen, Anna Goldie, Azalia Mirhoseini, Cameron McKinnon, et~al. 2022.
\newblock Constitutional ai: Harmlessness from ai feedback.
\newblock \emph{arXiv preprint arXiv:2212.08073}.

\bibitem[{Berant et~al.(2013)Berant, Chou, Frostig, and Liang}]{berant-etal-2013-semantic}
Jonathan Berant, Andrew Chou, Roy Frostig, and Percy Liang. 2013.
\newblock \href {https://www.aclweb.org/anthology/D13-1160} {Semantic parsing on {F}reebase from question-answer pairs}.
\newblock In \emph{Proceedings of the 2013 Conference on Empirical Methods in Natural Language Processing}, pages 1533--1544, Seattle, Washington, USA. Association for Computational Linguistics.

\bibitem[{Chen et~al.(2024)Chen, Li, Zhao, and Zhou}]{chen2024synthetic}
Hao Chen, Wenxin Li, Rui Zhao, and Yu~Zhou. 2024.
\newblock \href {https://arxiv.org/abs/2410.15226} {On the diversity of synthetic data and its impact on training large language models}.
\newblock \emph{arXiv preprint arXiv:2410.15226}.

\bibitem[{Chen et~al.(2021)Chen, Tworek, Jun, Yuan, Pinto, Kaplan, Edwards, Burda, Joseph, Brockman et~al.}]{chen2021evaluating}
Mark Chen, Jerry Tworek, Heewoo Jun, Qiming Yuan, Henrique Ponde De~Oliveira Pinto, Jared Kaplan, Harri Edwards, Yuri Burda, Nicholas Joseph, Greg Brockman, et~al. 2021.
\newblock Evaluating large language models trained on code.
\newblock \emph{arXiv preprint arXiv:2107.03374}.

\bibitem[{Cobbe et~al.(2021)Cobbe, Kosaraju, Bavarian, Chen, Jun, Kaiser, Plappert, Tworek, Hilton, Nakano et~al.}]{cobbe2021training}
Karl Cobbe, Vineet Kosaraju, Mohammad Bavarian, Mark Chen, Heewoo Jun, Lukasz Kaiser, Matthias Plappert, Jerry Tworek, Jacob Hilton, Reiichiro Nakano, et~al. 2021.
\newblock Training verifiers to solve math word problems.
\newblock \emph{arXiv preprint arXiv:2110.14168}.

\bibitem[{Dubey et~al.(2024)Dubey, Jauhri, Pandey, Kadian, Al-Dahle, Letman, Mathur, Schelten, Yang, Fan et~al.}]{dubey2024llama}
Abhimanyu Dubey, Abhinav Jauhri, Abhinav Pandey, Abhishek Kadian, Ahmad Al-Dahle, Aiesha Letman, Akhil Mathur, Alan Schelten, Amy Yang, Angela Fan, et~al. 2024.
\newblock The llama 3 herd of models.
\newblock \emph{arXiv preprint arXiv:2407.21783}.

\bibitem[{Dubois et~al.(2024)Dubois, Galambosi, Liang, and Hashimoto}]{dubois2024length}
Yann Dubois, Bal{\'a}zs Galambosi, Percy Liang, and Tatsunori~B Hashimoto. 2024.
\newblock Length-controlled alpacaeval: A simple way to debias automatic evaluators.
\newblock \emph{arXiv preprint arXiv:2404.04475}.

\bibitem[{Fan et~al.(2019)Fan, Jernite, Perez, Grangier, Weston, and Auli}]{fan-etal-2019-eli5}
Angela Fan, Yacine Jernite, Ethan Perez, David Grangier, Jason Weston, and Michael Auli. 2019.
\newblock \href {https://doi.org/10.18653/v1/P19-1346} {{ELI}5: Long form question answering}.
\newblock In \emph{Proceedings of the 57th Annual Meeting of the Association for Computational Linguistics}, pages 3558--3567, Florence, Italy. Association for Computational Linguistics.

\bibitem[{Fan et~al.(2018)Fan, Lewis, and Dauphin}]{fanHierarchical2018}
Angela Fan, Mike Lewis, and Yann Dauphin. 2018.
\newblock \href {https://arxiv.org/abs/1805.04833} {Hierarchical {{Neural Story Generation}}}.
\newblock \emph{Preprint}, arXiv:1805.04833.

\bibitem[{Gao et~al.(2024)Gao, Tow, Abbasi, Biderman, Black, DiPofi, Foster, Golding, Hsu, Le~Noac'h, Li, McDonell, Muennighoff, Ociepa, Phang, Reynolds, Schoelkopf, Skowron, Sutawika, Tang, Thite, Wang, Wang, and Zou}]{eval-harness}
Leo Gao, Jonathan Tow, Baber Abbasi, Stella Biderman, Sid Black, Anthony DiPofi, Charles Foster, Laurence Golding, Jeffrey Hsu, Alain Le~Noac'h, Haonan Li, Kyle McDonell, Niklas Muennighoff, Chris Ociepa, Jason Phang, Laria Reynolds, Hailey Schoelkopf, Aviya Skowron, Lintang Sutawika, Eric Tang, Anish Thite, Ben Wang, Kevin Wang, and Andy Zou. 2024.
\newblock \href {https://doi.org/10.5281/zenodo.12608602} {A framework for few-shot language model evaluation}.

\bibitem[{Ge et~al.(2024)Ge, Chan, Wang, Yu, Mi, and Yu}]{ge2024scaling}
Tao Ge, Xin Chan, Xiaoyang Wang, Dian Yu, Haitao Mi, and Dong Yu. 2024.
\newblock Scaling synthetic data creation with 1,000,000,000 personas.
\newblock \emph{arXiv preprint arXiv:2406.20094}.

\bibitem[{Grattafiori et~al.(2024)Grattafiori, Dubey, Jauhri, Pandey, Kadian, Al-Dahle, Letman, Mathur, Schelten, Vaughan et~al.}]{grattafiori2024llama}
Aaron Grattafiori, Abhimanyu Dubey, Abhinav Jauhri, Abhinav Pandey, Abhishek Kadian, Ahmad Al-Dahle, Aiesha Letman, Akhil Mathur, Alan Schelten, Alex Vaughan, et~al. 2024.
\newblock The llama 3 herd of models.
\newblock \emph{arXiv preprint arXiv:2407.21783}.

\bibitem[{Han et~al.(2022)Han, Kim, and Chang}]{han-etal-2022-measuring}
Seungju Han, Beomsu Kim, and Buru Chang. 2022.
\newblock \href {https://doi.org/10.18653/v1/2022.findings-emnlp.66} {Measuring and improving semantic diversity of dialogue generation}.
\newblock In \emph{Findings of the Association for Computational Linguistics: EMNLP 2022}, pages 934--950, Abu Dhabi, United Arab Emirates. Association for Computational Linguistics.

\bibitem[{Hendrycks et~al.(2020)Hendrycks, Burns, Basart, Zou, Mazeika, Song, and Steinhardt}]{hendrycks2020measuring}
Dan Hendrycks, Collin Burns, Steven Basart, Andy Zou, Mantas Mazeika, Dawn Song, and Jacob Steinhardt. 2020.
\newblock Measuring massive multitask language understanding.
\newblock \emph{arXiv preprint arXiv:2009.03300}.

\bibitem[{Jiang(2024)}]{jiang2024identifying}
Fengqing Jiang. 2024.
\newblock Identifying and mitigating vulnerabilities in llm-integrated applications.
\newblock Master's thesis, University of Washington.

\bibitem[{Kim et~al.(2024)Kim, Lee, Cho, Jang, Hwang, Won, Ahn, Lee, and Seo}]{kim2024knowledge}
Jiyeon Kim, Hyunji Lee, Hyowon Cho, Joel Jang, Hyeonbin Hwang, Seungpil Won, Youbin Ahn, Dohaeng Lee, and Minjoon Seo. 2024.
\newblock Knowledge entropy decay during language model pretraining hinders new knowledge acquisition.
\newblock \emph{arXiv preprint arXiv:2410.01380}.

\bibitem[{Kirk et~al.(2023)Kirk, Mediratta, Nalmpantis, Luketina, Hambro, Grefenstette, and Raileanu}]{kirk2023understanding}
Robert Kirk, Ishita Mediratta, Christoforos Nalmpantis, Jelena Luketina, Eric Hambro, Edward Grefenstette, and Roberta Raileanu. 2023.
\newblock Understanding the effects of rlhf on llm generalisation and diversity.
\newblock \emph{arXiv preprint arXiv:2310.06452}.

\bibitem[{Kwiatkowski et~al.(2019)Kwiatkowski, Palomaki, Redfield, Collins, Parikh, Alberti, Epstein, Polosukhin, Devlin, Lee et~al.}]{kwiatkowski2019natural}
Tom Kwiatkowski, Jennimaria Palomaki, Olivia Redfield, Michael Collins, Ankur Parikh, Chris Alberti, Danielle Epstein, Illia Polosukhin, Jacob Devlin, Kenton Lee, et~al. 2019.
\newblock Natural questions: a benchmark for question answering research.
\newblock \emph{Transactions of the Association for Computational Linguistics}, 7:453--466.

\bibitem[{Lambert et~al.(2024{\natexlab{a}})Lambert, Morrison, Pyatkin, Huang, Ivison, Brahman, Miranda, Liu, Dziri, Lyu, Gu, Malik, Graf, Hwang, Yang, Bras, Tafjord, Wilhelm, Soldaini, Smith, Wang, Dasigi, and Hajishirzi}]{lambert2024tulu3}
Nathan Lambert, Jacob Morrison, Valentina Pyatkin, Shengyi Huang, Hamish Ivison, Faeze Brahman, Lester James~V. Miranda, Alisa Liu, Nouha Dziri, Shane Lyu, Yuling Gu, Saumya Malik, Victoria Graf, Jena~D. Hwang, Jiangjiang Yang, Ronan~Le Bras, Oyvind Tafjord, Chris Wilhelm, Luca Soldaini, Noah~A. Smith, Yizhong Wang, Pradeep Dasigi, and Hannaneh Hajishirzi. 2024{\natexlab{a}}.
\newblock Tülu 3: Pushing frontiers in open language model post-training.

\bibitem[{Lambert et~al.(2024{\natexlab{b}})Lambert, Pyatkin, Morrison, Miranda, Lin, Chandu, Dziri, Kumar, Zick, Choi et~al.}]{lambert2024rewardbench}
Nathan Lambert, Valentina Pyatkin, Jacob Morrison, LJ~Miranda, Bill~Yuchen Lin, Khyathi Chandu, Nouha Dziri, Sachin Kumar, Tom Zick, Yejin Choi, et~al. 2024{\natexlab{b}}.
\newblock Rewardbench: Evaluating reward models for language modeling.
\newblock \emph{arXiv preprint arXiv:2403.13787}.

\bibitem[{Levesque et~al.(2012)Levesque, Davis, and Morgenstern}]{levesque2012winograd}
Hector Levesque, Ernest Davis, and Leora Morgenstern. 2012.
\newblock The winograd schema challenge.
\newblock In \emph{Thirteenth International Conference on the Principles of Knowledge Representation and Reasoning}. Citeseer.

\bibitem[{Li et~al.(2023)Li, Zhang, Dubois, Taori, Gulrajani, Guestrin, Liang, and Hashimoto}]{alpaca_eval}
Xuechen Li, Tianyi Zhang, Yann Dubois, Rohan Taori, Ishaan Gulrajani, Carlos Guestrin, Percy Liang, and Tatsunori~B. Hashimoto. 2023.
\newblock Alpacaeval: An automatic evaluator of instruction-following models.
\newblock \url{https://github.com/tatsu-lab/alpaca_eval}.

\bibitem[{Li et~al.(2024)Li, Chen, Xu, Qin, Xiao, Sun, and Luo}]{li2024entropic}
Ziniu Li, Congliang Chen, Tian Xu, Zeyu Qin, Jiancong Xiao, Ruoyu Sun, and Zhi-Quan Luo. 2024.
\newblock Entropic distribution matching in supervised fine-tuning of llms: Less overfitting and better diversity.
\newblock \emph{arXiv preprint arXiv:2408.16673}.

\bibitem[{Lin et~al.(2019)Lin, Zhou, Shen, Zhou, Bhagavatula, Choi, and Ren}]{lin2019commongen}
Bill~Yuchen Lin, Wangchunshu Zhou, Ming Shen, Pei Zhou, Chandra Bhagavatula, Yejin Choi, and Xiang Ren. 2019.
\newblock Commongen: A constrained text generation challenge for generative commonsense reasoning.
\newblock \emph{arXiv preprint arXiv:1911.03705}.

\bibitem[{McCoy et~al.(2023)McCoy, Yao, Friedman, Hardy, and Griffiths}]{mccoy2023embers}
R~Thomas McCoy, Shunyu Yao, Dan Friedman, Matthew Hardy, and Thomas~L Griffiths. 2023.
\newblock Embers of autoregression: Understanding large language models through the problem they are trained to solve.
\newblock \emph{arXiv preprint arXiv:2309.13638}.

\bibitem[{Mostafazadeh et~al.(2016)Mostafazadeh, Chambers, He, Parikh, Batra, Vanderwende, Kohli, and Allen}]{mostafazadeh2016corpus}
Nasrin Mostafazadeh, Nathanael Chambers, Xiaodong He, Devi Parikh, Dhruv Batra, Lucy Vanderwende, Pushmeet Kohli, and James Allen. 2016.
\newblock A corpus and evaluation framework for deeper understanding of commonsense stories.
\newblock \emph{arXiv preprint arXiv:1604.01696}.

\bibitem[{Mostafazadeh et~al.(2017)Mostafazadeh, Roth, Louis, Chambers, and Allen}]{mostafazadeh-etal-2017-lsdsem}
Nasrin Mostafazadeh, Michael Roth, Annie Louis, Nathanael Chambers, and James Allen. 2017.
\newblock \href {https://doi.org/10.18653/v1/W17-0906} {{LSDS}em 2017 shared task: The story cloze test}.
\newblock In \emph{Proceedings of the 2nd Workshop on Linking Models of Lexical, Sentential and Discourse-level Semantics}, pages 46--51, Valencia, Spain. Association for Computational Linguistics.

\bibitem[{O'Mahony et~al.(2024)O'Mahony, Grinsztajn, Schoelkopf, and Biderman}]{omahony2024attributing}
Laura O'Mahony, Leo Grinsztajn, Hailey Schoelkopf, and Stella Biderman. 2024.
\newblock Attributing mode collapse in the fine-tuning of large language models.
\newblock In \emph{ICLR 2024 Workshop on Mathematical and Empirical Understanding of Foundation Models}.

\bibitem[{Ouyang et~al.(2022)Ouyang, Wu, Jiang, Almeida, Wainwright, Mishkin, Zhang, Agarwal, Slama, Ray et~al.}]{ouyang2022training}
Long Ouyang, Jeffrey Wu, Xu~Jiang, Diogo Almeida, Carroll Wainwright, Pamela Mishkin, Chong Zhang, Sandhini Agarwal, Katarina Slama, Alex Ray, et~al. 2022.
\newblock Training language models to follow instructions with human feedback.
\newblock \emph{Advances in neural information processing systems}, 35:27730--27744.

\bibitem[{Qin et~al.(2024)Qin, Song, Hu, Yao, Cho, Wang, Wu, Liu, Liu, and Yu}]{qin2024infobench}
Yiwei Qin, Kaiqiang Song, Yebowen Hu, Wenlin Yao, Sangwoo Cho, Xiaoyang Wang, Xuansheng Wu, Fei Liu, Pengfei Liu, and Dong Yu. 2024.
\newblock Infobench: Evaluating instruction following ability in large language models.
\newblock \emph{arXiv preprint arXiv:2401.03601}.

\bibitem[{Reimers and Gurevych(2020)}]{reimers-2020-multilingual-sentence-bert}
Nils Reimers and Iryna Gurevych. 2020.
\newblock \href {https://arxiv.org/abs/2004.09813} {Making monolingual sentence embeddings multilingual using knowledge distillation}.
\newblock In \emph{Proceedings of the 2020 Conference on Empirical Methods in Natural Language Processing}. Association for Computational Linguistics.

\bibitem[{Research(2025)}]{thoughtworks2025semanticentropy}
Thoughtworks Research. 2025.
\newblock Evaluating llms using semantic entropy.
\newblock \url{https://www.thoughtworks.com/en-us/insights/blog/generative-ai/Evaluating-LLM-using-semantic-entropy}.
\newblock Accessed: 2025-05-18.

\bibitem[{Team(2024)}]{qwen2.5}
Qwen Team. 2024.
\newblock \href {https://qwenlm.github.io/blog/qwen2.5/} {Qwen2.5: A party of foundation models}.

\bibitem[{Team(2025)}]{qwen3}
Qwen Team. 2025.
\newblock \href {https://qwenlm.github.io/blog/qwen3/} {Qwen3: Think deeper, act faster}.

\bibitem[{Tevet and Berant(2021)}]{tevet-berant-2021-evaluating}
Guy Tevet and Jonathan Berant. 2021.
\newblock \href {https://doi.org/10.18653/v1/2021.eacl-main.25} {Evaluating the evaluation of diversity in natural language generation}.
\newblock In \emph{Proceedings of the 16th Conference of the European Chapter of the Association for Computational Linguistics: Main Volume}, pages 326--346, Online. Association for Computational Linguistics.

\bibitem[{Turpin et~al.(2023)Turpin, Michael, Perez, and Bowman}]{turpin2023language}
Miles Turpin, Julian Michael, Ethan Perez, and Samuel Bowman. 2023.
\newblock Language models don't always say what they think: Unfaithful explanations in chain-of-thought prompting.
\newblock \emph{Advances in Neural Information Processing Systems}, 36:74952--74965.

\bibitem[{Xia et~al.(2024)Xia, Xing, Du, Yang, Feng, Xu, Yin, and Xiong}]{xia2024fofo}
Congying Xia, Chen Xing, Jiangshu Du, Xinyi Yang, Yihao Feng, Ran Xu, Wenpeng Yin, and Caiming Xiong. 2024.
\newblock Fofo: A benchmark to evaluate llms' format-following capability.
\newblock \emph{arXiv preprint arXiv:2402.18667}.

\bibitem[{Yun et~al.(2025)Yun, Peng, and Shang}]{yun2025ultragen}
Longfei Yun, Letian Peng, and Jingbo Shang. 2025.
\newblock Ultragen: Extremely fine-grained controllable generation via attribute reconstruction and global preference optimization.
\newblock \emph{arXiv preprint arXiv:2502.12375}.

\bibitem[{Zhou et~al.(2023)Zhou, Lu, Mishra, Brahma, Basu, Luan, Zhou, and Hou}]{zhou2023instruction}
Jeffrey Zhou, Tianjian Lu, Swaroop Mishra, Siddhartha Brahma, Sujoy Basu, Yi~Luan, Denny Zhou, and Le~Hou. 2023.
\newblock Instruction-following evaluation for large language models.
\newblock \emph{arXiv preprint arXiv:2311.07911}.

\end{thebibliography}

\appendix

\newpage
\clearpage



\section{Fine-grained Ablation of Chat Template}
To investigate how structural prompting impacts output diversity in instruction-tuned language models, we adopt a stepwise ablation philosophy. The core idea is to isolate the individual contribution of special template tokens and dialog-style structure by incrementally removing formatting elements from the prompt. This approach allows us to disentangle whether diversity collapse is primarily driven by explicit tokens (e.g., \texttt{<|user|>}, \texttt{<s>}, [INST]) or by the broader conversational scaffold (e.g., system messages, turn markers).

We design four prompting modes with increasing degrees of simplification:
\begin{enumerate}
    \item full template: uses the model’s native chat format, including system headers, role markers, and delimiters;

    \item fake template: preserves structural layout but replaces special tokens with semantically meaningless placeholders, decoupling structure from token-level semantics;

    \item minimum.dialog: strips system messages and role tokens, retaining only natural language cues (e.g., user: / assistant:);

    \item simple steer: removes all structural elements, reducing the prompt to a bare instruction without any dialog framing. 
\end{enumerate}

All the prompts are shown in ~\autoref{tab:prompt_templates_all_models}.

\begin{table*}[htbp]
    \centering
    \scriptsize
    \renewcommand{\arraystretch}{1.5}
    \begin{tabularx}{\textwidth}{p{0.10\textwidth}|p{0.15\textwidth}|X}
        \hline
        \textbf{Model} & \textbf{Mode} & \textbf{Prompt} \\
        \hline
        \multirow{4}{*}{LLaMA}
        & full\_template & \texttt{<|begin\_of\_text|><|start\_header\_id|>user<|end\_header\_id|> Please write a news about a random topic.<|eot\_id|><|start\_header\_id|>assistant<|end\_header\_id|>} \\
        & fake\_template & \texttt{<\#init\_seq><@user\_name>user<@/user\_name> Please write a news about a random topic.<\#eot><@user\_name>assistant<@/user\_name>} \\
        & minimum\_dialog & \texttt{user: Please write a news about a random topic. \textbackslash n assistant:} \\
        & simple\_steer & \texttt{Please write a news about a random topic.} \\
        \hline
        \multirow{4}{*}{Qwen}
        & full\_template & \texttt{<|im\_start|>system You are Qwen, created by Alibaba Cloud. You are a helpful assistant.<|im\_end|>\textbackslash n<|im\_start|>user Please write a news about a random topic.<|im\_end|>\textbackslash n<|im\_start|>assistant} \\
        & fake\_template & \texttt{<\#meta\_start>sys You are Qwen, created by Alibaba Cloud. You are a helpful assistant.<\#meta\_end>\textbackslash n<\#meta\_start>usr Please write a news article about a random topic.<\#meta\_end>\textbackslash n<\#meta\_start>bot} \\
        & minimum\_dialog & \texttt{user: Please write a news about a random topic.\textbackslash n assistant:} \\
        & simple\_steer & \texttt{Please write a news about a random topic.} \\
        \hline
        \multirow{4}{*}{Tulu}
        & full\_template & \texttt{<|user|> Please write a news about a random topic. <|assistant|>} \\
        & fake\_template & \texttt{<<@@user@@>> Please write a news about a random topic. <<@@bot@@>>} \\
        & minimum\_dialog & \texttt{user: Please write a news about a random topic.\textbackslash n assistant:} \\
        & simple\_steer & \texttt{Please write a news about a random topic.} \\
        \hline
        \multirow{4}{*}{Mistral}
        & full\_template & \texttt{<s> [INST] Please write a news about a random topic. [/INST]} \\
        & fake\_template & \texttt{<@user> <Instruction> Please write a news about a random topic. </Instruction>} \\
        & minimum\_dialog & \texttt{user: Please write a news about a random topic.\textbackslash n assistant:} \\
        & simple\_steer & \texttt{Please write a news about a random topic.} \\
        \hline
        \multirow{4}{*}{Phi}
        & full\_template & \texttt{<|user|>\textbackslash nPlease write a news about a random topic.<|end|>\textbackslash n<|assistant|>\textbackslash n} \\
        & fake\_template & \texttt{<@user> Please write a news about a random topic. <@end> <@assistant>} \\
        & minimum\_dialog & \texttt{user: Please write a news about a random topic.\textbackslash n assistant:} \\
        & simple\_steer & \texttt{Please write a news about a random topic.} \\
        \hline
    \end{tabularx}
    \caption{
    Prompt templates used for different models and prompting modes. 
    Each row specifies how a model is instructed to generate a news article given the same semantic intent.
    While the wording remains constant across all conditions, variations in structural formatting (e.g., dialog tags, system headers, special tokens) reflect distinct learned priors for each model family.
    }
    \label{tab:prompt_templates_all_models}
\end{table*}

\section{Traditional Metrics}
\label{appendix:traditional_metrics}
Even when evaluated with traditional n-gram based diversity metrics instead of embedding-based semantic evaluations, we observe a consistent advantage in diversity for prompts without structured chat templates. As shown in ~\autoref{tab:traditional_metrics}, across all four Distinct-N scores (from Distinct-2 to Distinct-5), the Simple Steer prompt mode outperforms the Full Template for every model in the News Generation task. Likewise, self-BLEU, a metric that inversely reflects diversity (lower is better), is also consistently lower under Simple Steer. These results demonstrate that the observed diversity collapse is not an artifact of semantic embedding comparisons: even at the surface lexical level, the structured chat format severely restricts the model’s expressive variety, while a minimal steer prompt encourages broader lexical and topical generation.
\begin{table}[htbp]
    \centering
        \newcolumntype{g}{>{\columncolor{green!10}}c}
        \resizebox{0.48\textwidth}{!}{
            \begin{tabular}{llccccc}
            \toprule
            \multirow{2}*{Prompt Mode} & \multirow{2}*{Model} & 
            \multicolumn{5}{c}{\textbf{News Generation}} \\ 
            \cmidrule(lr){3-7} 
            & & Distinct-2~↑ & Distinct-3~↑ & Distinct-4~↑  & Distinct-5~↑ & self-BLEU~↓ \\
            \midrule
            Full Template & Llama-3-8B-Instruct  & 0.1556 & 0.3249 & 0.4699 & 0.5826 & 0.9319 \\
            \rowcolor{gray!20}
            Simple Steer & Llama-3-8B-Instruct  & \textbf{0.2107} & \textbf{0.4325} & \textbf{0.5971} & \textbf{0.7098} & \textbf{0.8884} \\
            Full Template & Tulu-3-8B-SFT  & 0.3646 & 0.6908 & 0.8615 & 0.9375 & 0.8186 \\
            \rowcolor{gray!20}
            Simple Steer & Tulu-3-8B-SFT  & \textbf{0.3987} & \textbf{0.7268} & \textbf{0.8834} & \textbf{0.9451} & \textbf{0.7884} \\
            Full Template & Qwen2.5-7B-Instruct  & 0.2158 & 0.4715 & 0.6532 & 0.7654 & 0.9157 \\
            \rowcolor{gray!20}
            Simple Steer & Qwen2.5-7B-Instruct  & \textbf{0.2469} & \textbf{0.5149} & \textbf{0.6940} & \textbf{0.8012} & \textbf{0.8908} \\
            Full Template & Mistral-7B-Instruct-v0.1  & 0.2192 & 0.4504 & 0.6208 & 0.7368 & 0.8969 \\
            \rowcolor{gray!20}
            Simple Steer & Mistral-7B-Instruct-v0.1  & \textbf{0.2657} & \textbf{0.5333} & \textbf{0.7066} & \textbf{0.8098} & \textbf{0.8599} \\
            Full Template & Phi-3.5-mini-instruct  & 0.2775 & 0.5943 & 0.7996 & 0.9030 & 0.8792 \\
            \rowcolor{gray!20}
            Simple Steer & Phi-3.5-mini-instruct  & \textbf{0.3515} & \textbf{0.6887} & \textbf{0.8630} & \textbf{0.9384} & \textbf{0.8351} \\
            \bottomrule
            \end{tabular}
        }
        \caption{News generation diversity scores (Distinct-N~↑ and self-BLEU~↓) for different prompt modes and models. In each pair, the better metric is bolded (higher for Distinct-N, lower for self-BLEU).}
        \label{tab:traditional_metrics}
\end{table}

\section{Downstream Evaluation Details}
\label{appendix:downstream_evaluation_details}

We follow the standardized evaluation setups provided by the \texttt{lm-evaluation-harness} framework ~\citep{eval-harness} for all downstream tasks. Below, we outline the key configurations for each benchmark:

\begin{enumerate}
\item \textbf{GSM8K:} We use the gsm8k/main dataset in free-form generation mode (generate\_until) with a deterministic decoding setting (temperature 0.0). The model is prompted with five few-shot examples (num\_fewshot=5), and predictions are evaluated using an exact match metric after applying a flexible-extract filter to extract the final numerical answer.

\item \textbf{MMLU:} We include four subject groups under the mmlu group (mmlu\_stem, mmlu\_other, mmlu\_social\_sciences, mmlu\_humanities) and compute accuracy (acc) as the evaluation metric. Final performance is aggregated by dataset size to reflect a balanced view across subjects.

\item \textbf{HumanEval}: We evaluate the model's ability to generate correct Python code using the openai/openai\_humaneval dataset. The generation is truncated on common code delimiters (e.g., \texttt{\textbackslash n class}, \texttt{\textbackslash n def}) and evaluated with the \texttt{pass@1} metric, which measures the fraction of problems solved correctly on the first attempt. We use the canonical check(entry\_point) setup as the target for correctness evaluation.

\item \textbf{Web QS}: We evaluate open-domain factual QA using the web\_questions dataset, following the multiple-choice evaluation protocol. Each question is formatted as Question: \texttt{<question>\textbackslash n Answer}:, and the model selects one answer from a predefined list of candidates. The metric used is exact match, which checks whether the predicted answer exactly matches any of the ground-truth answers. Aggregation is performed using the mean over all test examples. We enable decontamination filtering by matching questions against known training data to avoid data leakage. This setup follows the v2.0 configuration of the evaluation suite.

\item \textbf{IFEval:} We evaluate instruction-following capabilities using the IFEval benchmark (google/IFEval). Each example consists of a prompt designed to assess compliance with specific instructions. The task is configured in generate\_until mode with deterministic decoding (temperature = 0.0), and we use zero-shot prompting (num\_fewshot = 0). Evaluation is performed using one accuracy-based metrics:

\begin{enumerate}
    \item Instance-level loose accuracy: Aggregated instance-level score under the relaxed matching criterion.
\end{enumerate}

\item \textbf{WSC273:} We evaluate commonsense reasoning and coreference resolution using the wsc273 subset of the Winograd Schema Challenge. Each input consists of a sentence with an ambiguous pronoun that must be resolved to the correct antecedent. The model selects from two choices, which are formed by substituting each candidate into the sentence prefix up to the pronoun location. The task is evaluated using multiple choice format, and accuracy (acc) is used as the evaluation metric. Final results are computed as the mean accuracy over all test examples. We apply decontamination by checking for overlaps with the original sentence text to prevent potential data leakage. This configuration follows version 1.0 of the benchmark.

\end{enumerate}

All tasks use the latest available version from the benchmark suite, and configurations are aligned with prior work to ensure comparability and reproducibility.

\end{document}